\documentclass{article}

\usepackage[numbers]{natbib}
\usepackage[final]{nips_2017}
\usepackage[utf8]{inputenc} 
\usepackage[T1]{fontenc}    
\usepackage{hyperref}       
\usepackage{url}            
\usepackage{booktabs}       
\usepackage{amsfonts}       
\usepackage{nicefrac}       
\usepackage{microtype}      

\usepackage{tikz} 
\usepackage{amsthm} 
\usepackage{mathtools} 
\usepackage[skip=0pt]{subcaption} 
\usepackage{algorithm} 
\usepackage{algpseudocode} 
\usepackage{multirow} 
\theoremstyle{definition}
\newtheorem{definition}{Definition}
\newtheorem{lem}{Lemma}
\usepackage{amsmath}
\usepackage{amssymb}
\usepackage{thmtools}
\usepackage{bm} 
\usepackage{textcomp}

\DeclareMathOperator*{\argmax}{\arg\!\max}

\declaretheorem{theorem}
\declaretheoremstyle[%
  spaceabove=-6pt,%
  spacebelow=6pt,%
  headfont=\normalfont\itshape,%
  postheadspace=1em,%
  qed=\qedsymbol%
]{mystyle} 
\declaretheorem[name={Proof sketch},style=mystyle,unnumbered,
]{prf}

\declaretheoremstyle[%
  spaceabove=-6pt,%
  spacebelow=6pt,%
  headfont=\normalfont\itshape,%
  postheadspace=1em,%
  qed=\qedsymbol%
]{mystyle} 
\declaretheorem[name={Proof},style=mystyle,unnumbered,
]{prf2}

\title{On the Real-Time Vehicle Placement Problem}

\author{
  Abhinav Jauhri
  \qquad
  Carlee Joe-Wong
  \qquad
  John Paul Shen\\ 
  Electrical \& Computer Engineering Department \\
  Carnegie Mellon University \\
  Moffett Field, CA 94035\\
  \texttt{\{ajauhri,cjoewong,jpshen\}@andrew.cmu.edu}
}

\begin{document}
\maketitle

\begin{abstract}
Motivated by ride-sharing platforms' efforts to reduce their riders' wait times for a vehicle, this paper introduces a novel problem of placing vehicles to fulfill real-time pickup requests in a spatially and temporally changing environment. The real-time nature of this problem makes it fundamentally different from other placement and scheduling problems, as it requires not only real-time placement decisions but also handling real-time request dynamics, which are influenced by human mobility patterns. We use a dataset of ten million ride requests from four major U.S. cities to show that the requests exhibit significant self-similarity. We then propose distributed online learning algorithms for the real-time vehicle placement problem and bound their expected performance under this observed self-similarity.
\end{abstract}

\section{Introduction}
In the past five years, ride-sharing platforms like Uber and Lyft have become a significant means of transportation, accounting for nearly two billion rides in 2016. 
Given this popularity, ride-sharing companies have turned their attention towards the problem of optimizing rider experience. In particular, they have begun to examine ways to minimize rider wait times. Future vehicular technologies like autonomous cars can also benefit from algorithms to reduce wait times.

Reducing wait times to below two minutes is a challenging problem that requires real-time vehicle placement. The real-time nature of this problem then introduces two challenges: first, drivers should proactively predict where future pickup requests will be located and go to these locations in anticipation of future pickup requests, eliminating passenger waiting due to vehicle travel time. Yet to accurately predict future requests in the next few minutes, drivers must account for not just overall patterns in ride requests, but also real-time temporal and geo-spatial request fluctuations, which are influenced by human mobility dynamics. Second, these vehicle (driver) placement decisions must be made quickly, which precludes any significant coordination between different vehicles. 

Existing work in this area has tended to focus on a coarser version of the problem that does not address one or both of these two challenges, e.g., \cite{miao2016taxi} examines the placement of taxis based on predicted demand at a much coarser geo-spatial scale. The problem of load balancing by distributing vehicles around a geographical space is studied by~\cite{miao2017data}; the authors state that making ride request predictions to balance load at a small time granularity (less than thirty minutes) makes the problem hard. There has also been work using offline learning methods to optimally match riders and drivers under the assumption that all travel plans are known in advance \cite{jia2016optimization}, and \cite{qu2014cost} provides a recursion tree approach to maximize driver profit by recommending routes that are likely to have multiple pickup points. Probabilistic reasoning was applied by~\cite{ziebart2008navigate} to provide routing based on taxi drivers' collective intelligence. However, these latter approaches do not deliver real-time placements.

In this paper, we formulate the  fine-grained (at a timescale of less than a few minutes and spatial scale of a few hundred meters) real time vehicle placement  problem, and propose a distributed online learning approach to this problem.
An online learning algorithm gives us the flexibility to consider 
real-time ride request patterns in making our placement decisions. Historical information can help inform these decisions, but the utility of such historical information may vary 
a lot geo-spatially. 
We are not aware of any attempts to use online learning algorithms for real-time vehicle placement. Crucially, 
we discover a pattern which is exploited to analyze the performance of our proposed online real-time vehicle placement algorithms. 



\section{Problem Definition}
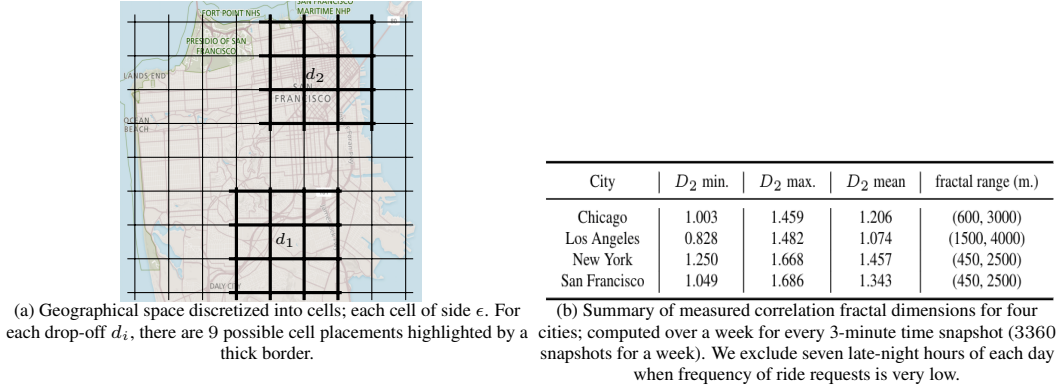
\begin{figure}
\footnotesize
    \captionsetup[sub]{font=scriptsize,justification=centering}
    \begin{subfigure}[t]{.5\textwidth}
        \centering
        \begin{tikzpicture}[remember picture, overlay]
            \node[inner sep=0pt] at (2,2) {\includegraphics[width=4cm,height=4cm]{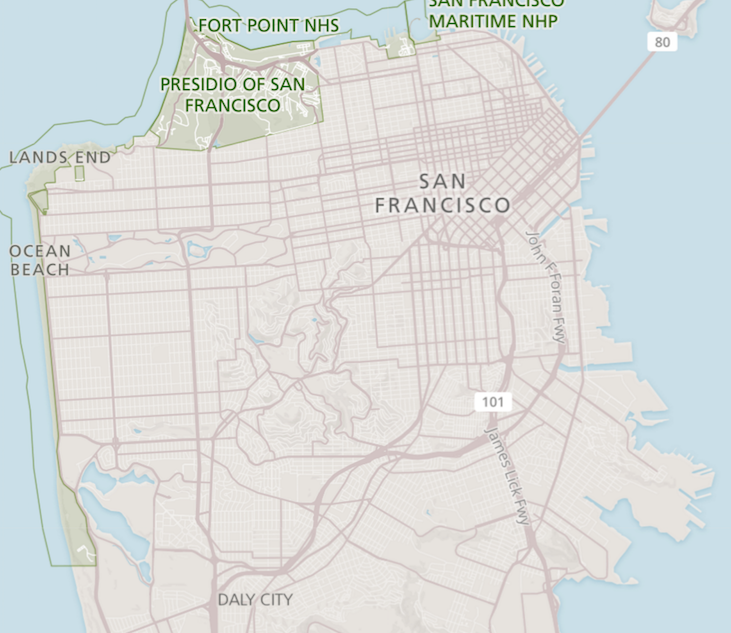}};
        \end{tikzpicture}
        \begin{tikzpicture}
            \node at (0.2,-1.1) {\tiny $d_1$};
            \node at (0.6,1.1) {\tiny $d_2$};
            \draw[step=.45cm] (-1.9,-1.9) grid (1.9,1.9);
            \draw[step=.45cm, very thick] (-0.55,-1.9) grid (.95,-.4);
            \draw[step=.45cm, very thick] (-0.15,0.35) grid (1.4,1.9);
        \end{tikzpicture}
        \caption{Geographical space discretized into 
        cells; each cell of side $\epsilon$. For each drop-off $d_i$, there are $9$ possible cell placements highlighted by a thick border.}
        \label{fig:gridded_surface}
    \end{subfigure}%
    \captionsetup[sub]{font=scriptsize,justification=centering}
    \begin{subfigure}[t]{.5\textwidth}
        \includegraphics[width=7cm, height=2cm]{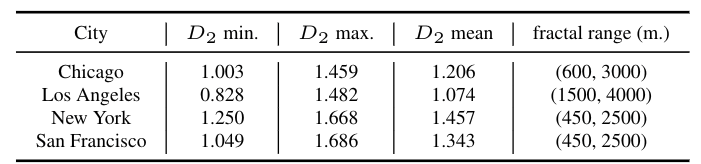}
        \caption{Summary of measured correlation fractal dimensions for four cities; computed over a week for every 3-minute time snapshot ($3360$ snapshots for a week). We exclude seven late-night hours of each day  when frequency of ride requests is very low.}
        \label{tab:fractal_dimensions}
    \end{subfigure}
    \caption{Illustration of vehicle pick-ups and drop-offs and an observed $D_2$ values.}
\end{figure}

We consider a geographical surface that is divided into equally sized cells of side length $\epsilon$ each, as shown in Figure~\ref{fig:gridded_surface}, to get an aggregate of $a\times b$ cells. We suppose a consecutive set of time snapshots, taken at intervals $\tau_\epsilon$, where each time snapshot $t$ specifies a set of drop-offs at different cells, represented by the matrix $\boldsymbol D_t \in \mathbb Z^{a \times b}$, and pickups $\boldsymbol P_t \in \mathbb Z^{a \times b}$ of ride requests. Each $(i,j)$ entry in $\boldsymbol D_t$ or $\boldsymbol P_t$ is the sum of all drop-offs or pickups that occurred at $t$, in cell $(i,j)$. In this work, we have kept the length of
each time snapshot interval to be a few minutes, reflecting our goal of keeping rider wait times under two minutes. The \textit{real time vehicle placement} problem is to decide how to move vehicles from drop-off cells at time snapshot $t$ to neighboring cells such that pickups happen at those cells at $t+1$. Formally, our goal is to place vehicles in the ``best'' neighboring cells, i.e., so as to maximize a reward at time snapshot $t$ that is defined as:
\begin{equation}
R_t(\boldsymbol P_t, \boldsymbol \Gamma_t) = \frac{1}{n_t}\sum_{i,j}
min(\boldsymbol P_t[i,j], \boldsymbol \Gamma_t[i,j])
\label{eq:reward}
\end{equation}
where: $\boldsymbol \Gamma_t \in \mathbb Z^{a \times b}$; each entry represents the number of vehicles placed at each cell, or zero otherwise. $n_t = \sum_i^a \sum_j^b \boldsymbol D_{t-1}[i,j] = 
\sum_i^a \sum_j^b \boldsymbol \Gamma_{t}[i,j] $; the number of drop-offs at $t-1$ or equivalently, the number of placements at $t$. For any $t$, $\boldsymbol P_t$ is known and the goal is to find $\boldsymbol \Gamma_t$ using $\boldsymbol D_{t-1}$ such that $R_t$ is maximized. The reward $R_t$ thus represents the average number of successful pick-ups per cell. 


The algorithms decide the 
placement of each vehicle without knowing where future pick-ups will occur.
Moreover, for every drop-off by a vehicle at time $t-1$ at cell $(i,j)$, its 
corresponding placement at $t$ may be either $(i,j)$ itself or one of 
the neighbouring cells. This constraint is shown by the bolded cell outlines in Figure~\ref{fig:gridded_surface}. The number of neighbouring cells at which the vehicle can be placed is determined by the vehicle's ability to travel to the neighbouring cell within the time snapshot interval; it is thus determined by the length of the 
snapshot interval ($\tau_{\epsilon}$) and $\epsilon$. This constraint also ensures that vehicles incur negligible cost (e.g., gasoline used) in moving to the exact pick-up location, as they only move to neighbouring cells. We denote the radius of the permitted neighbourhood by $\epsilon'$. The goal of our online algorithm is then to choose the vehicle placements within any time interval $[t - 1,t)$ so as to maximize the reward (\ref{eq:reward}), subject to the constraint that vehicles cannot move too far from the cell where they dropped off passengers at time $t - 1$. Any excess placements for $t$ are not considered in the reward function for pickups after $t$; for $t+1$, placements are computed using drop-offs at $t$ only. 

The real time vehicle placement problem generalizes other known problems in computer science. For instance, in the $k$-server problem one must place $k$ servers within a metric space so as to fulfill requests that can come in at any point in the space~\citep{manasse1990competitive}. Viewing the servers as vehicles, our problem is a dynamic version of the $k$-server problem: we must adjust the vehicle placements in real time as new requests arrive, subject to constraints on the travel time of the vehicles. 

\section{Fractals and Ride Requests}
In this section, we provide some background~\citep{belussi1998estimating} on fractals, or self-similarity, and show that our ride request data exhibits self-similar patterns. It has been observed that real datasets of point-sets representing Montgomery county, Long Beach county, and CA road intersections exhibit properties of self-similarity~\citep{proietti1999complexity}. In our problem we are also dealing with a natural phenomenon involving human behaviors, i.e. human mobility in urban environments. Characterizing human mobility patterns using fractals provides a succinct description which could be leveraged in a variety of applications and their analyses, including our vehicle placement problem.

A set of points in some high dimensional space is a fractal if it exhibits self-similarity over all or a range of scales. This can be illustrated by a \textit{Sierpinski triangle} which is formed by recursively removing a smaller triangle contained within an equilateral triangle, and repeating the process for all smaller triangles. The result are smaller  triangles which look like replicas of the whole triangle. The characteristic property of fractals is this self-similarity property: smaller parts of fractal are similar to the whole fractal. After numerous recursive iterations of a Sierpinski triangle, it becomes evident that it is neither a 1-dimensional or 2-dimensional Euclidean object~\citep{belussi1998estimating}. Instead, we characterize these shapes with \textit{fractal dimensions}. The fractal dimension also indicates whether point-sets exhibit self-similarity.

\subsection{Fractal Dimension}
Fractal dimension provides a quantitative measure of self-similarity for a point-set. In the vehicle placement problem, the reward achieved by any algorithm depends on the underlying spatial patterns of pick-ups and drop-offs, which, as we observe, exhibits self-similarity. We shall focus on one specific measure of fractal dimension which relates to the estimation of spatial queries like how many points are present in a cell. Points in our context represent pickup or drop-off locations. Ride requests exhibit a biased distribution, i.e., they are more probable in high-density areas. 

Consider a high dimensional space divided into (hyper-)cubic grid cells of side $\epsilon$. Let $p_i$ denote the occupancy of the $i-$th cell which is the number of points which fall into the $i-$th cell of the grid.

\begin{definition}{\textit{Correlation fractal dimension:}}
For a point-set that has self-similarity in the fractal range $(\epsilon_1, \epsilon_2)$ the correlation fractal dimension $D_2$ is defined as:
\begin{equation}
D_2 \equiv \frac{\partial \log \sum_i p_i^2}{\partial \log \epsilon} 
= \text{constant} \qquad \epsilon \in (\epsilon_1, \epsilon_2)
\end{equation}
\end{definition}


For a point-set to be self-similar, in the range $(\epsilon_1, \epsilon_2)$, we observe the plot of $\log \sum_i p_i^2$ versus $\log \epsilon$ must be a straight line, implying self-similarity in that range, i.e. the fractal range~\citep{belussi1998estimating}. The slope of the line is equal to the correlation fractal dimension $D_2$.

\textbf{Self Similarity in Ride Requests:} Our dataset is a collection of real ride requests in a city over one week~\citep{jauhri2017space}. Each data item represents a ride request, i.e., a 4-tuple of: 1) time of request; 2) pickup location (latitude \& longitude); 3) drop-off location (latitude \& longitude); 4) time of drop-off.

To compute the correlation fractal dimension ($D_2$), our point-set comprises of pick-up locations of ride requests. We discretize time into short snapshots of 3-minute durations and divide the geographical space into square cells each with side $\epsilon = 100$ meters. Figure~\ref{tab:fractal_dimensions} presents the correlation fractal dimensions, $D_2$, from the ride request patterns of the four cities. For each city, there is a consistent weekly pattern of ride requests and a corresponding variation of its $D_2$ values. For each city, the values of $D_2$ range between $D_2$min and $D_2$max. The mean values for the four cities range from 1.0 to 1.5, indicating the degree of self-similarity of ride request patterns varies from city to city. The fractal range also varies across the cities; however, other than Los Angeles, the other cities all exhibit very similar fractal ranges, from 500 meters to 2500 meters. 

The fractal dimension and the fractal range succinctly capture the characteristics of ride request patterns in a city. We believe these fractal dimensions and associated fractal ranges can be instrumental in revealing effective solutions to the real-time vehicle placement problem and facilitating the assessment of the effectiveness of specific algorithms for this problem.    

\section{Potential Algorithms}\label{sec:algorithms}
In this section, we examine in detail three potential algorithms for the real-time vehicle placement problem. We make use of the fractal analysis in the previous section to analyze their effectiveness. For each drop-off, the possible set of neighboring cells to which the vehicle can potentially be moved are restricted; see Figure~\ref{fig:gridded_surface}. If the drop-off location is in the $(i,j)$ cell, then the set of neighboring cells within a square-shaped neighborhood centered at $(i,j)$ and containing $c^2$ cells is denoted as $\eta(\square, (i,j), c)$. Figure~\ref{fig:gridded_surface} illustrates neighborhoods with $c^2$ = 3x3 cells.

\textbf{Baseline Algorithm:} Our simplest baseline (Algorithm~\ref{alg:uniform_random}) assumes a uniform weight across all neighbouring cells without considering any past history on the number of requests in those cells. 
\begin{algorithm}
\scriptsize
\begin{algorithmic}[1]
\Require Search radius $\epsilon'$; cell length $\epsilon$; $\boldsymbol D_i, \boldsymbol P_i~\forall i \in \{1,\ldots, k\}$.
\For{each time snapshot $t = 1,2,..., k$ }
\State $\boldsymbol \Gamma_{t+1} = 0, \in \mathbb R^{a\times b}$
\For{each non-zero entry $(i,j) \in \boldsymbol D_t $}
\State Pick a set of cells $N = \eta (\square, (i,j), \frac{2\epsilon'}{\epsilon})$.
\State Choose a cell $\in N$ uniformly at random 
with coordinates $(i', j')$.
\State $\boldsymbol \Gamma_{t+1}[i', j'] = \boldsymbol \Gamma_{t+1}[i',j'] + 1$
\EndFor
\State Obtain rewards for $R_{t+1}(\boldsymbol P_{t+1}, \boldsymbol \Gamma_{t+1})$
\EndFor
\end{algorithmic}
\caption{URand-NH (Uniform Random with No History)}
\label{alg:uniform_random}
\end{algorithm}

\textbf{Poisson Process Based Algorithm:} To capture the temporal pattern of ride requests, we consider a placement approach that models ride requests for every grid as a Poisson Process, as used in~\citep{zhang2014dmodel} (Algorithm~\ref{alg:pp_lh}). Numerous works~\citep{castellano2009statistical, oliveira2005human} have recommended against modeling human behavior as a Poisson process, due to the assumption made in the Poisson Process that consecutive events happen at regular time intervals. Contrarily, human activity tends to be bursty for short time intervals accompanied by long intervals of inactivity~\citep{barabasi2005origin, alfi2007conference}. For this reason, our maximum likelihood estimator (MLE) estimator is based on limited, and most recent history of inter arrival times of ride requests. Formally, let $N(t), t \geq 0 $ be the number of ride requests (events) that occurred by time $t$.
\begin{lem}\label{lem:pp}
The probability of at least one event occurring in the time snapshot $t$ is given by:
\begin{equation}\label{eq:pp}
\Pr\{N(t) > 0 | \lambda\} = 1 - e^{-\lambda t}
\end{equation}
\begin{prf2}
See~\citep{harchol2013performance}.
\end{prf2}
\end{lem}

\begin{algorithm}
\scriptsize
\begin{algorithmic}[1]
\Require Search radius $\epsilon'$; cell length $\epsilon$; $\boldsymbol D_i, \boldsymbol P_i \forall i \in \{1,..., k\}$; history length $m$; min. samples $u$.
\For{each time snapshot $t = (m+1), (m+2), ..., k$ }
\State $\boldsymbol M = 0 \in \mathbb R^{a \times b}$
\State $\boldsymbol \Gamma_{t+1} = 0, \in \mathbb R^{a\times b}$
\State $\boldsymbol M = \sum_{i=t-m}^m \boldsymbol D_i + \boldsymbol P_i$
\State Estimate $\boldsymbol \lambda$ for each cell entry having a value $>u$ in $\boldsymbol M$ by computing MLE using inter-arrival times.
\For{each non-zero entry $(i,j) \in \boldsymbol D_t $}
\State Pick a set of cells $N = \eta (\square, (i,j), \frac{2\epsilon'}{\epsilon})$.
\State Choose a cell $(i',j') = \argmax_{(i,j) \in N} \Pr\{N(1) > 0| \boldsymbol \lambda[i,j]\}$ (see Equation~\ref{eq:pp}).
\State $\boldsymbol \Gamma_{t+1}[i', j'] = \boldsymbol \Gamma_{t+1}[i',j'] + 1$
\State $\boldsymbol M[i', j'] = 0$ \Comment{Not discarding $(i',j')$, degrades performance since it might be considered again.}
\EndFor
\State Obtain rewards for $R_{t+1}(\boldsymbol P_{t+1}, \boldsymbol \Gamma_{t+1})$
\EndFor
\end{algorithmic}
\caption{PP-LH (Poisson Process with Limited History)}
\label{alg:pp_lh}
\end{algorithm}

For any instance when none of the neighbourhood cells have an estimator $\in \boldsymbol \lambda$, the algorithm falls back to uniform random strategy (like for URand-NH) from within the set of neighbors. 

\textbf{Follow The Leader (FTL) Algorithm:} Follow The Leader is a sequential prediction strategy, which always puts all the weight on the cell with the highest reward so far. It has been applied to multiple online problems for time series predictions~\citep{de2014follow}. If there is a tie among the leaders, i.e., there are multiple leaders in the neighbourhood with equal reward values, a leader is chosen uniformly at random. If all cells are zero, FTL-CH also falls back to a uniform random strategy.
\begin{algorithm}
\scriptsize
\begin{algorithmic}[1]
\Require Search radius $\epsilon'$; cell length $\epsilon$ $\boldsymbol D, \boldsymbol P$; initial history length $m$.
\For{each time snapshot $t = (m+1), (m+2), ..., k$ }
\State $\boldsymbol M = 0 \in \mathbb R^{a \times b}$
\State $\boldsymbol \Gamma_{t+1} = 0, \in \mathbb R^{a\times b}$
\State $\boldsymbol M = \sum_{i=1}^{t-1} \boldsymbol D_i + \boldsymbol P_i$
\For{each non-zero entry $(i,j) \in \boldsymbol D_t $}
\State Pick a set of cells $N = \eta (\square, (i,j), \frac{2\epsilon'}{\epsilon})$.
\State Choose a cell $(i',j') = \argmax_{(i,j) \in N} \boldsymbol M[i, j]$. 
\State $\boldsymbol \Gamma_{t+1}[i', j'] = \boldsymbol \Gamma_{t+1}[i',j'] + 1$
\State $\boldsymbol M[i', j'] = \boldsymbol M[i', j'] - 1 $ \Comment{Update to history improves performance.}
\EndFor
\State Obtain rewards for $R_{t+1}(\boldsymbol P_{t+1}, \boldsymbol \Gamma_{t+1})$
\EndFor
\end{algorithmic}
\caption{FTL-CH (Follow The Leader with Complete History)}
\label{alg:ftl_ch}
\end{algorithm}

\section{Analysis \& Results}\label{sec:analysis}
In this section we analyze the performance of our proposed algorithms. We use the fact that our ride request data exhibits self-similarity to provide an average bound on the algorithms' reward functions.

\textbf{Reward Analysis:} Given the observation of self-similarity, we are primarily interested in approximating queries such as: what is the average number of neighboring cells which are intersected by some neighborhood shape (like square or circle) of a certain length. Let $\overline{nb}(\epsilon')$ denote the average number of points within an enclosed square $\square$ of radius $\epsilon'$ where $\epsilon' \in (\epsilon_1, \epsilon_2)$ for which self-similarity is observed. Note $\epsilon' > \epsilon$. An important consequence of this is: 

\begin{lem}\label{lem:neighs}
Given a set of points $\mathcal{P}$ with finite cardinality and its Correlation Dimension $D_2$, the average number of points within the shape $\square$ with radius of $\epsilon'$ follows the power law: $\overline{nb}(\epsilon') \propto \epsilon'^{D_2}$.
\end{lem}
\begin{prf2}
See~\citep{belussi1998estimating}.
\end{prf2}

We now evaluate the performance of our algorithms under this self-similarity assumption. We assume throughout that our performance is determined by the placement strategy, i.e., the number of drop-offs is smaller than the number of pick-up requests in a given neighbourhood. Otherwise, if the number of drop-offs is too large, any strategy is likely to perform well. We show in our numerical analysis (Section 5.2) that this assumption holds in our dataset.
\begin{theorem}\label{th:1}
Suppose that the total number of drop-offs in a given cell of radius $\epsilon'$ is no more than the expected number of pick-up requests in a cell. Then the expected performance of FTL-CH is strictly better than URand-NH for any pickup matrix $\boldsymbol P_t$ with $1 < D_2 < 2$: $\mathbb{E}_{\text{FTL-CH}}[R_t] > \mathbb{E}_{\text{URand-NH}}[R_t]$,
and the expected number of fulfilled pick-ups per drop-off with FTL-CH exceeds that for URand-NH.
\end{theorem}

\begin{prf} In a given grid with radius $\epsilon'$, the expected reward for the Follow The Leader with Complete History (FTL-CH) algorithm at time $t$ is given by the minimum of the number of drop-offs in that grid at time $t$, which we denote by $D$, and the number of pickups at time $t + 1$ within the cell $(i,j)$ chosen by the FTL-CH algorithm: all drop-offs at time $t$ are placed in cell $(i,j)$ at time $t + 1$. The expected number of pickups in this cell is then proportional to $\epsilon^{D_2}$ from Lemma~\ref{lem:neighs}, and the expected reward is $\min\left\{D,C\epsilon^{D_2}\right\}$ for some constant $C > 0$.

If $D \leq C\epsilon^{D_2}$, then the expected reward with FTL-CH is $D$, which is also the maximum possible reward. Thus, the uniform random strategy cannot perform any better.

To show the second part of the theorem, we note that for the uniform random strategy, the expected fraction of total pick-ups in the grid that can be fulfilled by one drop-off is $\epsilon^2/{\epsilon'}^2$, i.e., the reciprocal of the number of cells in a grid. Then the expected number of pickups fulfilled is the expected total number of pickups multiplied by this fraction, i.e., $C{\epsilon'}^{D_2 - 2}\epsilon^2$ = $C\epsilon^{D_2}\left(\epsilon/\epsilon'\right)^{2 - D_2} \leq C\epsilon^{D_2}$, the expected number of pickups under FTL-CH, since $1 < D_2 < 2$ and $\epsilon \leq \epsilon'$.
%
\end{prf}

\begin{theorem}\label{th:2}
Suppose that the expected number of future pickup requests in each cell at time $t$ is given by the number of requests experienced in the past interval of $[t-1, t)$, and the interarrival times of a sequence of pickups follow an Exponential distribution. Then the expected performance of FTL-CH is equivalent to that of PP-LH: $\mathbb{E}_{\text{FTL-CH}}[R_t] \equiv \mathbb{E}_{\text{PP-LH}}[R_t]$.
\end{theorem}
\begin{prf}
Using Lemma~\ref{lem:pp} with $t=1$, we claim that the placement cell chosen by PP-LH will on average have experienced higher pick-ups than any of its neighbouring cells for a fixed time interval. This is true because with the interval $[t-1,t)$, the maximum likelihood estimator for the Exponential distribution is the inverse of the mean inter-arrival times. The resulting $\lambda$ estimate will be large for cells which have high rate of events, or equivalently cells which observed more events in the given time interval. From (\ref{eq:pp}), we then see that PP-LH will choose the cell with the largest $\lambda$ estimate, i.e., the cell that experienced the most number of events. Thus, the average reward obtained by PP-LH will be equivalent to that of FTL-CH~\footnote{For a more detailed proof refer to Lemma 1 in \url{https://classes.soe.ucsc.edu/cmps290c/Spring09/lect/10/wmkalai-rewrite.pdf}.}.
\end{prf}





\begin{figure*}[!t]
\captionsetup{justification=centering}
\centering
\begin{subfigure}{.5\textwidth}
  \centering
    \includegraphics[height=1.1in,width=\textwidth]{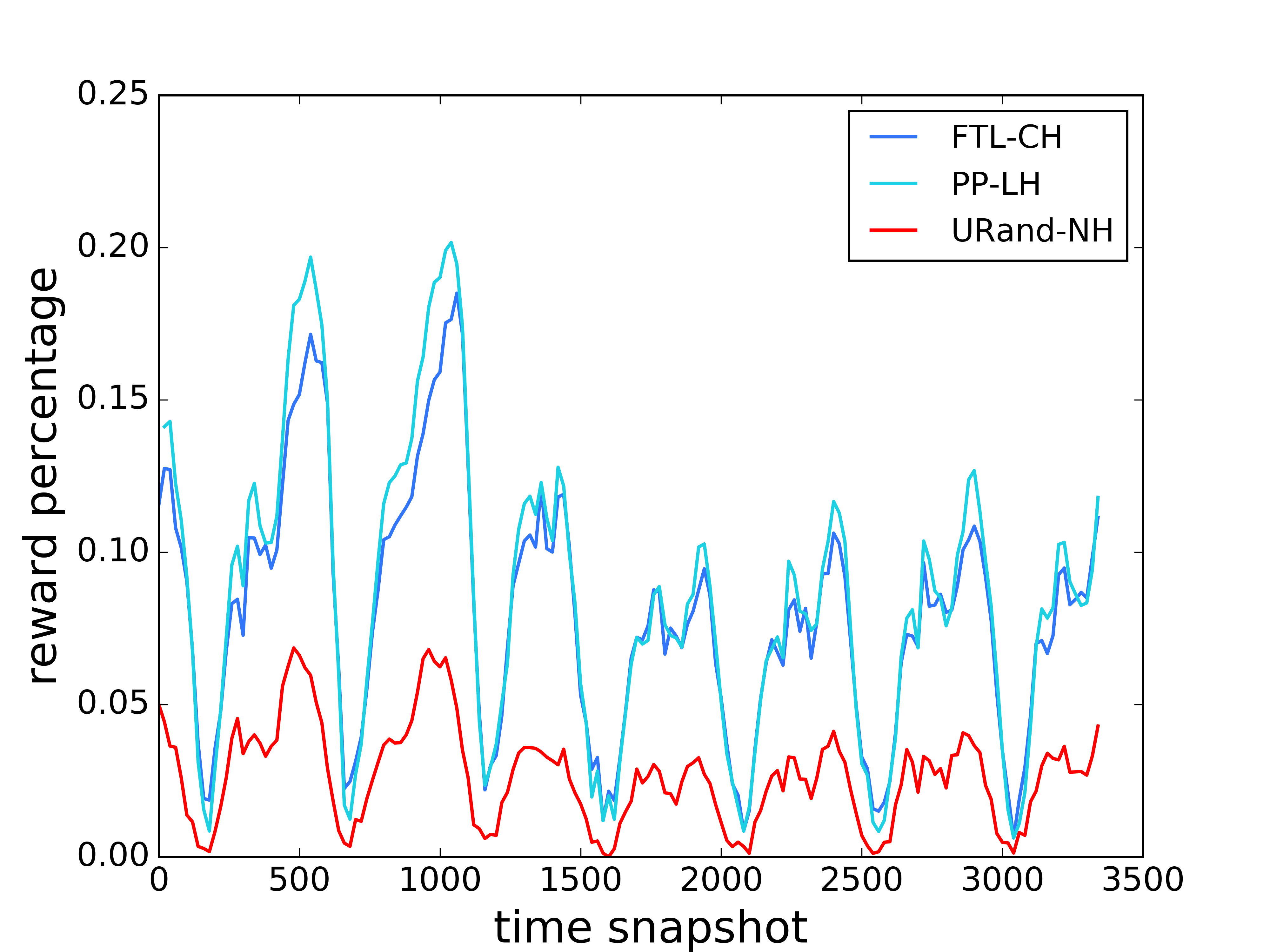}
    \label{fig:reward_chicago}
    \caption{Chicago}
\end{subfigure}%
\begin{subfigure}{.5\textwidth}
  \centering
    \includegraphics[height=1.1in,width=\textwidth]{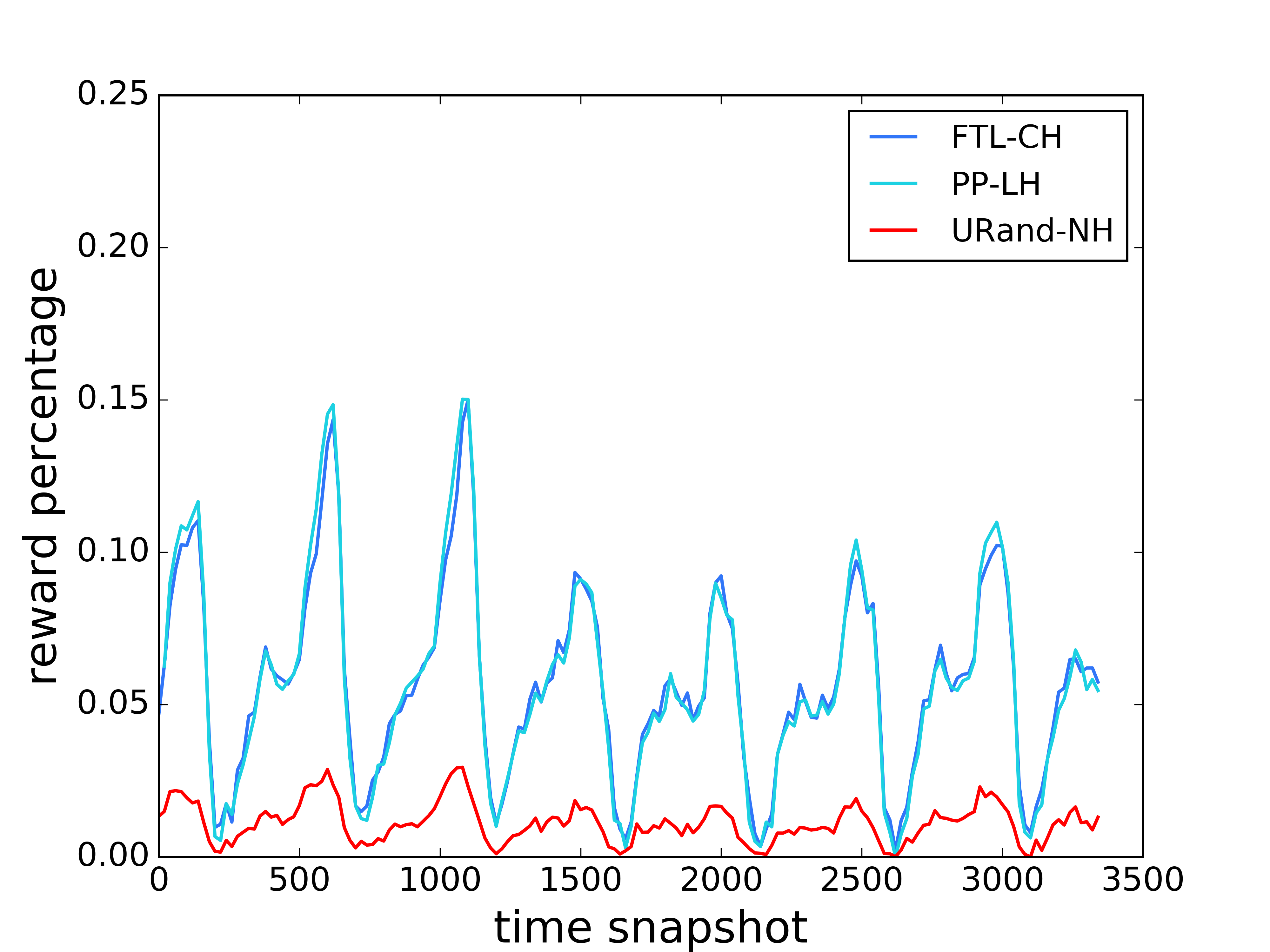}
    \label{fig:reward_los_angeles}
    \caption{Los Angeles}
\end{subfigure}
\begin{subfigure}{.5\textwidth}
  \centering
    \includegraphics[height=1.1in,width=\textwidth]{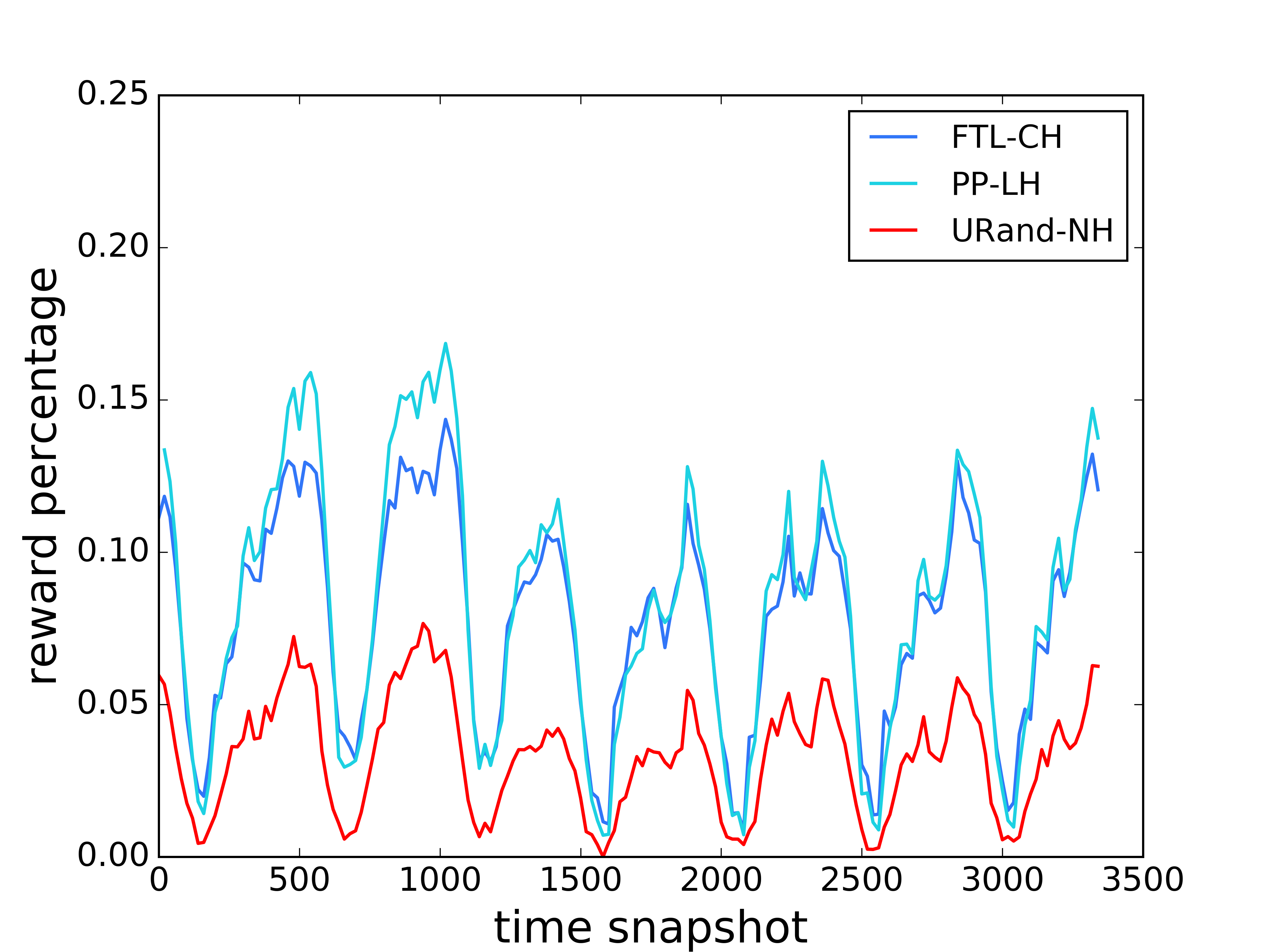}
    \label{fig:reward_new_york}
    \caption{New York}
\end{subfigure}%
\begin{subfigure}{.5\textwidth}
  \centering
    \includegraphics[height=1.1in,width=\textwidth]{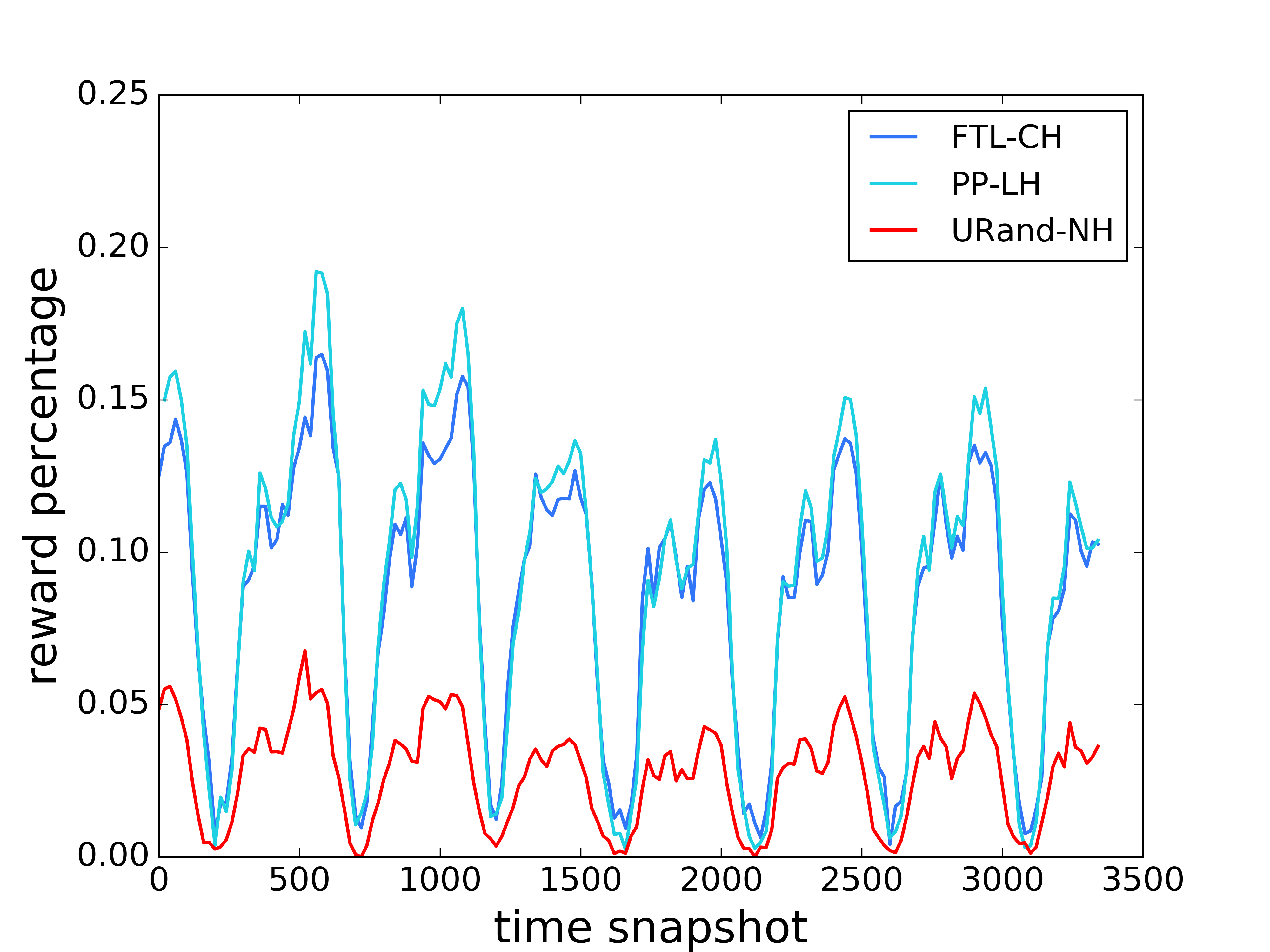}
    \label{fig:reward_san_francisco}
    \caption{San Francisco}
\end{subfigure}
\caption{The PP-LH algorithm out-performs FTL-CH slightly and URand-NH significantly across all four cities in terms of the reward (Eq.~\ref{eq:reward}) for a week with three minute time snapshots.}
\label{fig:reward_plots}
\end{figure*}

\textbf{Experimental Results:} We apply the three algorithms to the ride request data sets from four cities and find the achieved rewards in each, indicating the effectiveness of the algorithms.
Figure~\ref{fig:reward_plots} plots the reward functions for the three algorithms for each 3-minute ($\tau_{\epsilon}$) time interval over a week. $\epsilon'=500$ meters which effectively means there are $10^2$ possible cells (total cells $= (\frac{2\epsilon'}{\epsilon})^2$) for any vehicle to be placed in. In Algorithms~\ref{alg:pp_lh} and \ref{alg:ftl_ch}, $m=20$ and $3$ respectively.

The weekly 7-day pattern of these functions can be clearly seen. All three algorithms tend to be more effective when there are large numbers of ride requests. Supporting the claims in Theorems~\ref{th:1} and \ref{th:2}, PP-LH and FTL-CH perform similarly for all four cities. On average, the reward percentage for PP-LH is $0.75\%$ better than FTL-CH. In San Francisco, PP-LH guarantees that $11.7\%$ of pickups occur almost instantaneously, roughly within thirty seconds of the ride requests, since our cells are of length $\epsilon=100$ meters. The average rewards using PP-LH for Chicago, Los Angeles, and New York yield similar results of $11.1\%, 7.2\%, ~\&~10.2\%$ respectively. Los Angeles has the lowest expected reward as well as the lowest average correlation fractal dimension $D_2$ (Figure~\ref{tab:fractal_dimensions}).


\begin{figure}[!t]
\captionsetup{justification=centering}
\centering
\begin{subfigure}{.5\textwidth}
  \centering
    \includegraphics[height=1.1in,width=\textwidth]{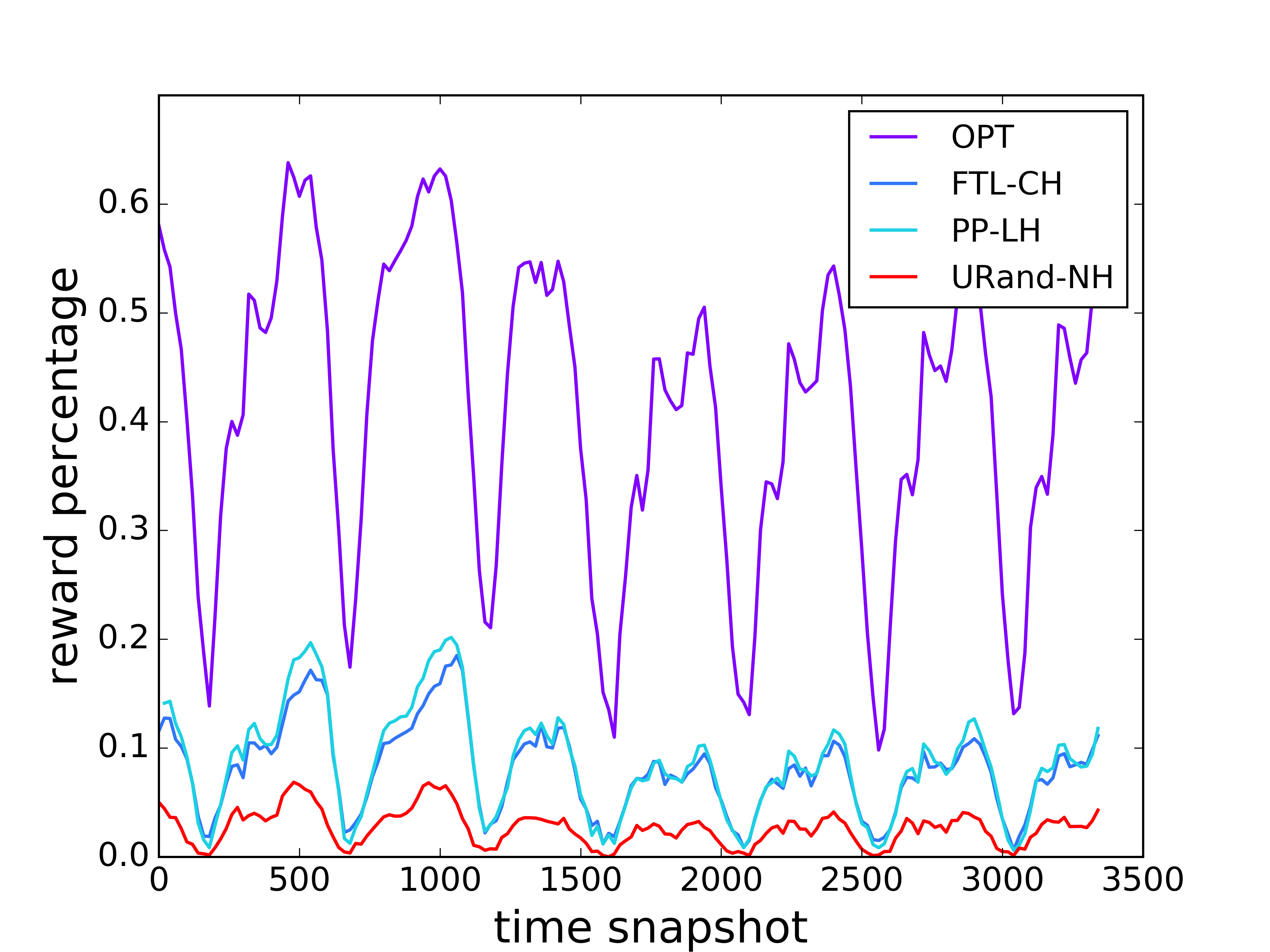}
    \label{fig:reward_chicago_opt}
    \caption{Chicago}
\end{subfigure}%
\begin{subfigure}{.5\textwidth}
  \centering
    \includegraphics[height=1.1in,width=\textwidth]{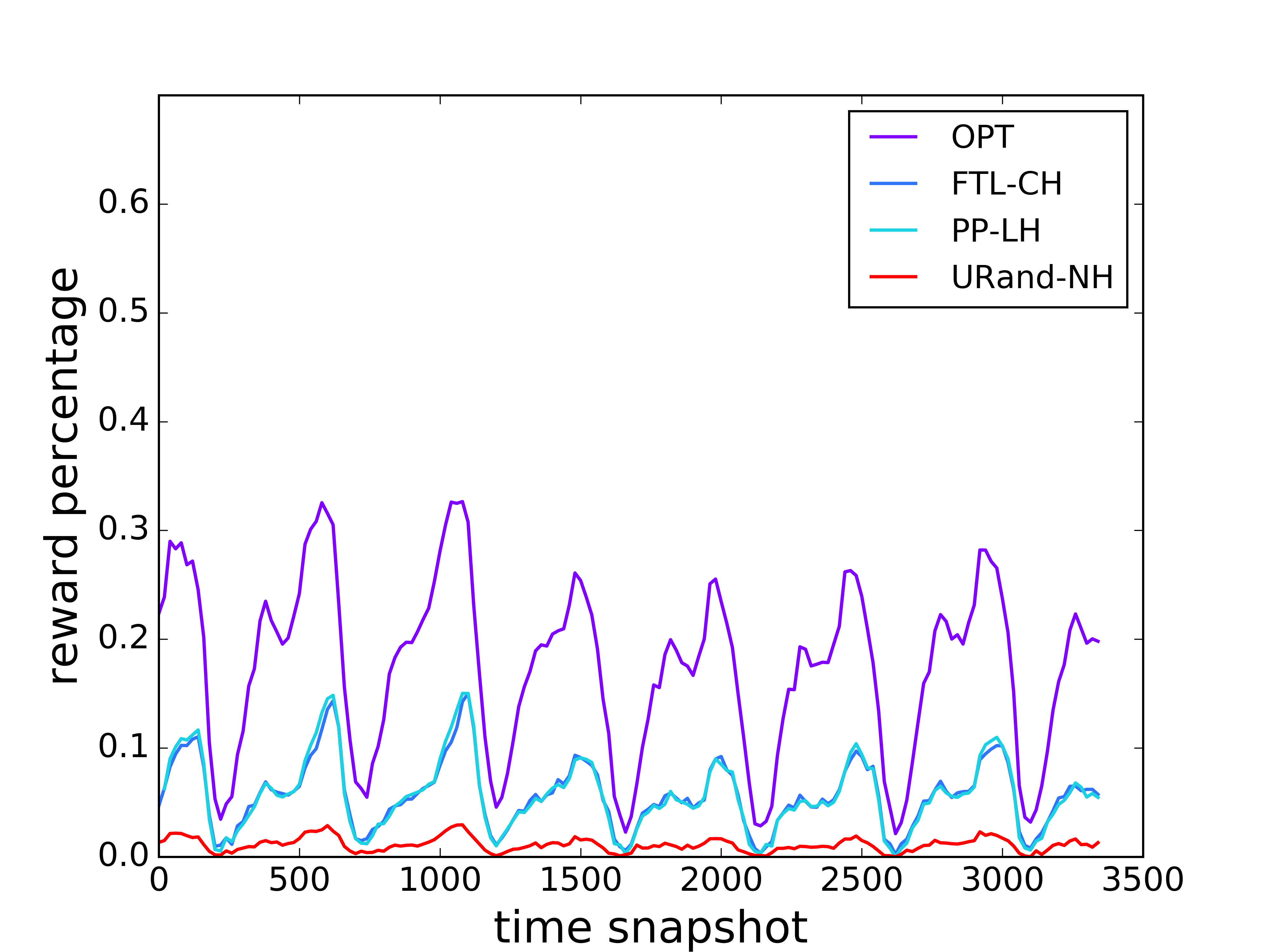}
    \label{fig:reward_los_angeles_opt}
    \caption{Los Angeles}
\end{subfigure}
\begin{subfigure}{.5\textwidth}
  \centering
    \includegraphics[height=1.1in,width=\textwidth]{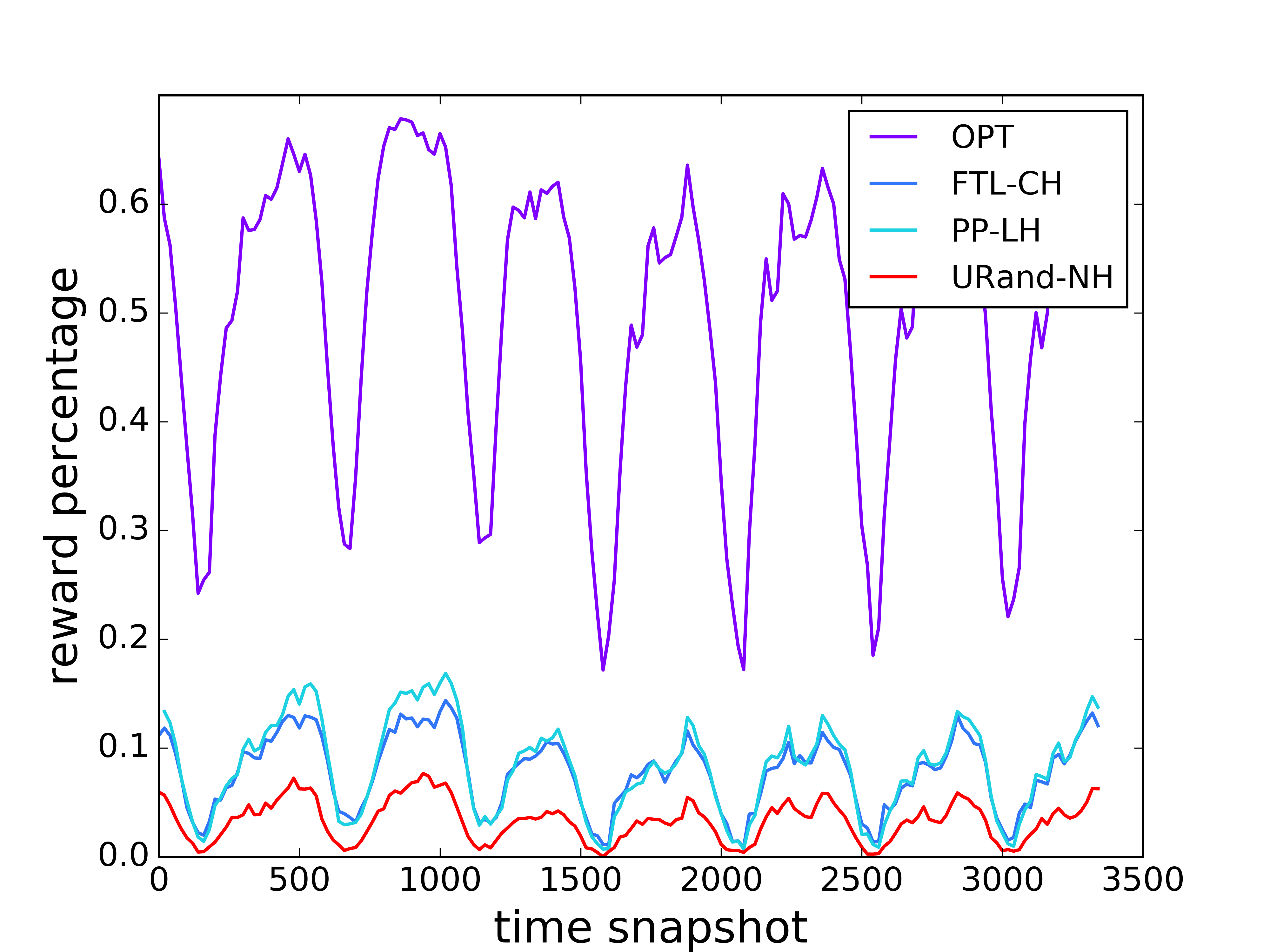}
    \label{fig:reward_new_york_opt}
    \caption{New York}
\end{subfigure}%
\begin{subfigure}{.5\textwidth}
  \centering
    \includegraphics[height=1.1in,width=\textwidth]{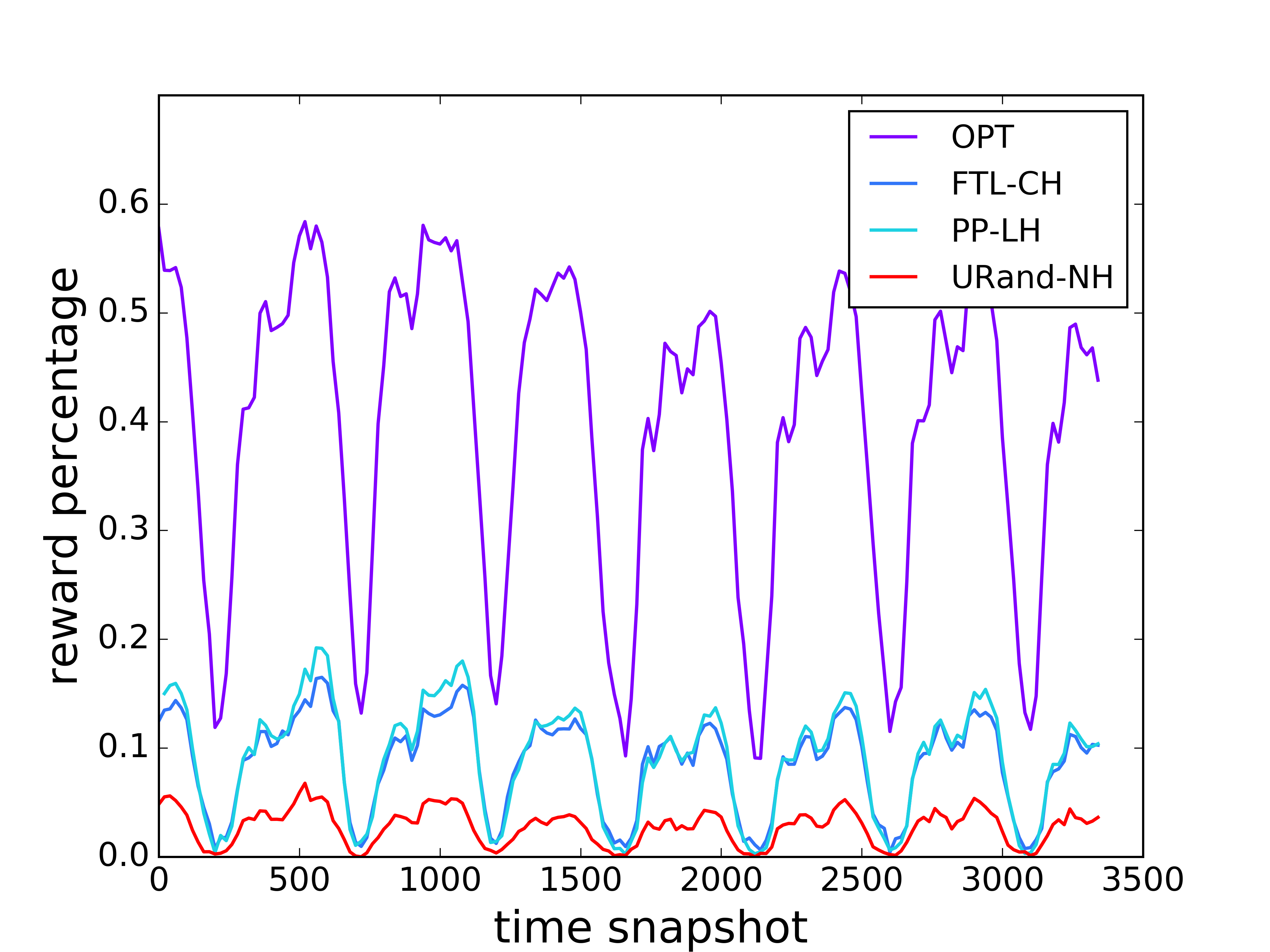}
    \label{fig:reward_san_francisco_opt}
    \caption{San Francisco}
\end{subfigure}
\caption{Comparison of reward percentage plots for 3 algorithms along with optimal (OPT) reward.}
\label{fig:reward_plots_opt}
\end{figure}

\textbf{Discussion \& Future Work:} We considered three existing algorithms for the real-time vehicle placement problem. Even though PP-LH performed better than the other two algorithms, in comparison to OPT, as shown in Figure~\ref{fig:reward_plots_opt}, there is still much room for further improvement. The reward function for OPT is computed using actual data; for every drop-off OPT assumes perfect knowledge of future pickups in the neighbouring cells.


There are several important insights from this work. 1) History of past pickups and drop-offs helps; although when accompanied with some sort of randomization does not necessarily help to boost the performance of the algorithm. For instance, uniform random selection of the leader as in the case of our FTL-CH, does as good as selecting a leader with the lowest index~\cite{blum2007learning}. 2) We believe randomization is essential but some weighting mechanism needs to be considered which can be characterized with the help of self-similarity. 3) Having limited most recent history is almost as good as having complete history. One obvious extension is to consider vehicle placement actions not just in the next time snapshot but on a longer time scale.

\textbf{Acknowledgements:} This project is funded in part by Carnegie Mellon University\textquotesingle s National USDOT University Transportation Center for Mobility, Mobility21, which is sponsored by the US Department of Transportation. 
We also thank referees for their helpful comments. 
\bibliography{refs}
\bibliographystyle{plain}
\end{document}